# Deep Reinforcement Learning Designed Shinnar-Le Roux RF Pulse using Root-Flipping: DeepRF$_{SLR}$

Dongmyung Shin, Sooyeon Ji, Doohee Lee, Jieun Lee, Se-Hong Oh, and Jongho Lee

*Abstract*—A novel approach of applying deep reinforcement learning to an RF pulse design is introduced. This method, which is referred to as DeepRF$_{SLR}$, is designed to minimize the peak amplitude or, equivalently, minimize the pulse duration of a multiband refocusing pulse generated by the Shinar Le-Roux (SLR) algorithm. In the method, the root pattern of SLR polynomial, which determines the RF pulse shape, is optimized by iterative applications of deep reinforcement learning and greedy tree search. When tested for the designs of the multiband pulses with three and seven slices, DeepRF$_{SLR}$ demonstrated improved performance compared to conventional methods, generating shorter duration RF pulses in shorter computational time. In the experiments, the RF pulse from DeepRF$_{SLR}$ produced a slice profile similar to the minimum-phase SLR RF pulse and the profiles matched to that of the computer simulation. Our approach suggests a new way of designing an RF by applying a machine learning algorithm, demonstrating a "machine-designed" MRI sequence.

## I. Introduction

In MRI, deep learning has been widely applied not only in image processing but also in image formation such as image reconstruction, parametric mapping, and artifact correction [1]-[6]. Most of these works utilized supervised learning that pairs an input (e.g., aliased image) with the desired output (e.g., aliasing free image) to train a deep neural network (DNN). Another type of deep learning is deep reinforcement learning [7]. In this approach, a DNN is trained to perform a series of actions in a given environment to maximize a reward. One example of the applications of deep reinforcement learning is to learn the Breakout game [8]. Initially, a DNN agent randomly moves the control stick, resulting in a poor score. With increasing numbers of trials, the agent learns a policy that maximizes the game score. The policy is updated in each game and, therefore, no explicitly labeled data are necessary, making reinforcement learning different from supervised learning. Recently, deep reinforcement learning has been applied to MRI applications such as encoding gradient generation [9], view plane search [10], and k-space sampling optimization [11], [12].

An RF pulse plays a key role in controlling spin magnetization in MRI. The design of an RF pulse requires careful considerations for shape, duration, slice profile, peak RF amplitude, specific absorption rate (SAR), etc. Several design methods such as the filter-design algorithm [13], optimal control theory [14], and mathematical transformation [15] have been proposed. Deep learning may provide an effective solution in generating an RF pulse. So far, however, only a few studies utilized (shallow) neural networks for the design of an RF pulse [16]-[18]: Gezelter *et al*. obtained Fourier coefficients of an RF pulse for a desired slice profile using a single hidden layer network [16]; Mirfin *et al*. applied a single hidden layer network to design a parallel transmission RF pulse [17]; Vinding *et al*. suggested a multi-dimensional RF pulse design using a four-layer neural network [18].

Recently, multiband RF pulses, which concurrently excite multiple slices to speed up data acquisition, have been developed [19]. The design of a multiband RF pulse is often limited by SAR and/or peak RF amplitude. To address the peak RF amplitude limitation, a few methods have been developed [20]-[22]. In particular, the method proposed by Sharma *et al*. suggested an approach that simplifies the design by reformulating a multiband spin-echo pulse design to a problem of finding the optimal binary pattern using a Monte-Carlo algorithm, resulting in a reduced peak RF amplitude [21].

In this study, we present a novel approach that utilizes deep reinforcement learning to design a multiband RF pulse with a reduced peak amplitude or, equivalently, a shorter RF duration. We demonstrate that deep reinforcement learning combined with a greedy tree search can efficiently optimize a multiband RF pulse even for a large number of bands (NB) and time-bandwidth product (TBW). This new method is referred to as DeepRF$_{SLR}$ hereafter. The source code of DeepRF$_{SLR}$ is available at https://github.com/SNU-LIST/DeepRF_SLR.

This work was supported by the National Research Foundation of Korea grant (NRF-2018R1A2B3008445), Samsung Research Funding & Incubation Center (SRFC-IT1801-09), and Samsung Electronics.
  D. Shin, S. Ji, D. Lee, J. Lee, and J. Lee are with the Department of Electrical and Computer Engineering, Seoul National University, Seoul 08826, Republic of Korea. (e-mail: shinsae11@gmail.com).
  S-H. Oh is with the Division of Biomedical Engineering, Hankuk University of Foreign Studies, Gyeonggi-do 17035, Republic of Korea.

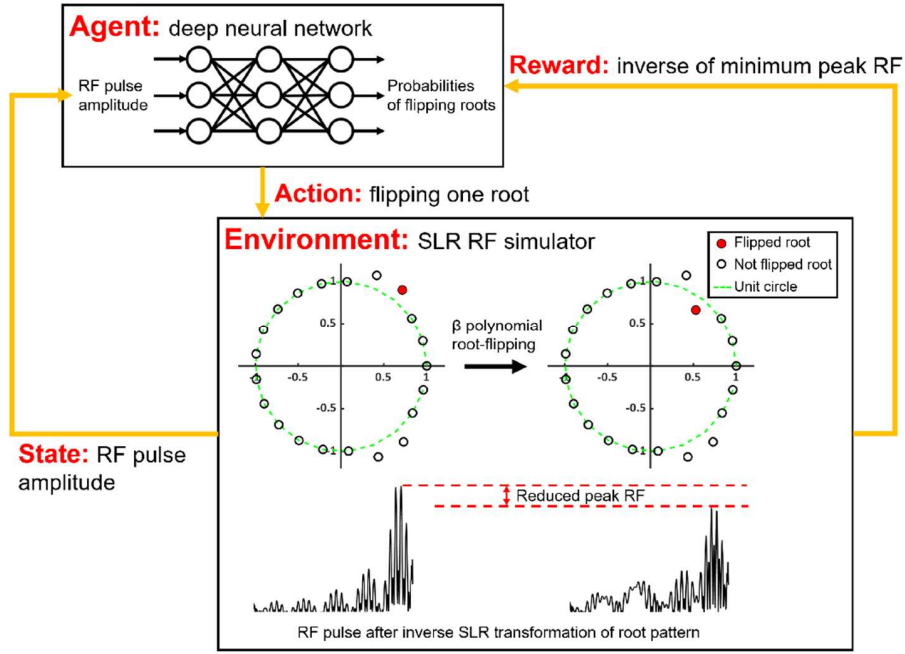

Fig. 1. Primary components of deep reinforcement learning in DeepRF$_{SLR}$. The agent, a deep neural network, takes the RF pulse amplitude (state) as an input and outputs probabilities of flipping the roots. The action is to flip one of the roots based on the probabilities given by the agent. The reward is the inverse of the minimum peak RF. The environment is an SLR RF simulator.

## II. METHODS

### A. Root-flipping in Shinnar-Le Roux RF Design

DeepRF$_{SLR}$ adopts the Shinnar-Le Roux (SLR) RF pulse design method, which converts an RF pulse design into the generation of two complex polynomials denoted as α and β [13]. These polynomials are determined by a filter design algorithm for a target slice profile. Once an SLR RF pulse is designed, the amplitude of the RF can be further modified while maintaining the magnitude slice profile by flipping the roots of the polynomials [23], [24]. This "root-flipping" can impose a necessary characteristic for RF (e.g., minimum-phase RF) and has been applied to minimize the peak RF amplitude for single-band inversion, saturation, and refocusing pulses [25]-[27]. Recently, studies by Sharma *et al*. [21] and Seada *et al*. [28] extended the root-flipping approach for a multiband spin-echo RF pulse design. To search root patterns, Sharma *et al*. [21] utilized the Monte-Carlo algorithm whereas Seada *et al*. [28] applied the genetic algorithm, demonstrating a substantial reduction in peak RF. However, these search algorithms may not be effective in finding the optimal root pattern for a large number of roots.

### B. DeepRF$_{SLR}$

The purpose of DeepRF$_{SLR}$ is to perform an efficient search of the root patterns for a multiband refocusing RF pulse, which can have a large number of roots. To achieve this goal, DeepRF$_{SLR}$ combines deep reinforcement learning with a greedy tree search algorithm [29]. The primary components of deep reinforcement learning for DeepRF$_{SLR}$ are defined as follows (Fig. 1). The state is the amplitude or envelope of an RF. The agent is a DNN that generates the probabilities of flipping the roots for a given state. The action is to flip one root using the output of the agent. The reward is defined as the inverse of the minimum peak RF. The environment is an SLR RF simulator that transforms a root pattern into an RF pulse [13].

We formulate DeepRF$_{SLR}$ as an episodic task in deep reinforcement learning [7]. The process of one episode is summarized in Fig. 2a. Initially, the DNN takes the amplitude of the minimum-phase RF pulse [21] as an input and produces the probabilities of flipping for eligible roots (see II.C for the definition of eligible roots). With the flipping chance of each root being proportional to the given probability, one of the roots is flipped stochastically. Then, the new root pattern is inverse SLR transformed to generate a new RF [13]. After that, the amplitude of this RF is fed to the input of the DNN (Fig. 2a), repeating the procedure $N_{root}$ times in one episode where $N_{root}$ is the number of the roots eligible for flipping. At the end of each episode, a reward is calculated as the inverse of the minimum peak RF among the $N_{root}$ RFs generated during the episode. Then, the neural network weights are updated using the policy gradient method [30], [31] (Fig. 2a).

After updating the network weights, the greedy tree search starts from the root pattern that has the minimum peak RF in the episode (Fig. 2b). For the greedy tree search, each eligible root is flipped, generating distinct $N_{root}$ root patterns that have only one different root compared to the original root pattern. Then, the tree grows with the pattern of the minimum peak as the starting point of the next search until no lower peak RF is generated. A greedy choice, which has the smallest peak amplitude, is selected as the result of the greedy tree search. This result is saved, and the algorithm restarts the deep reinforcement learning stage (Fig. 2c). DeepRF$_{SLR}$ iterates the deep reinforcement learning and greedy tree search (i.e., whole process of Fig. 2) until the number of flipping reaches a pre-defined value. Finally, once the algorithm stops, the best solution among the saved results is chosen as the final design. Note that the learning phase of DeepRF$_{SLR}$ happens as an on-going process and, therefore, no separate learning session is required.

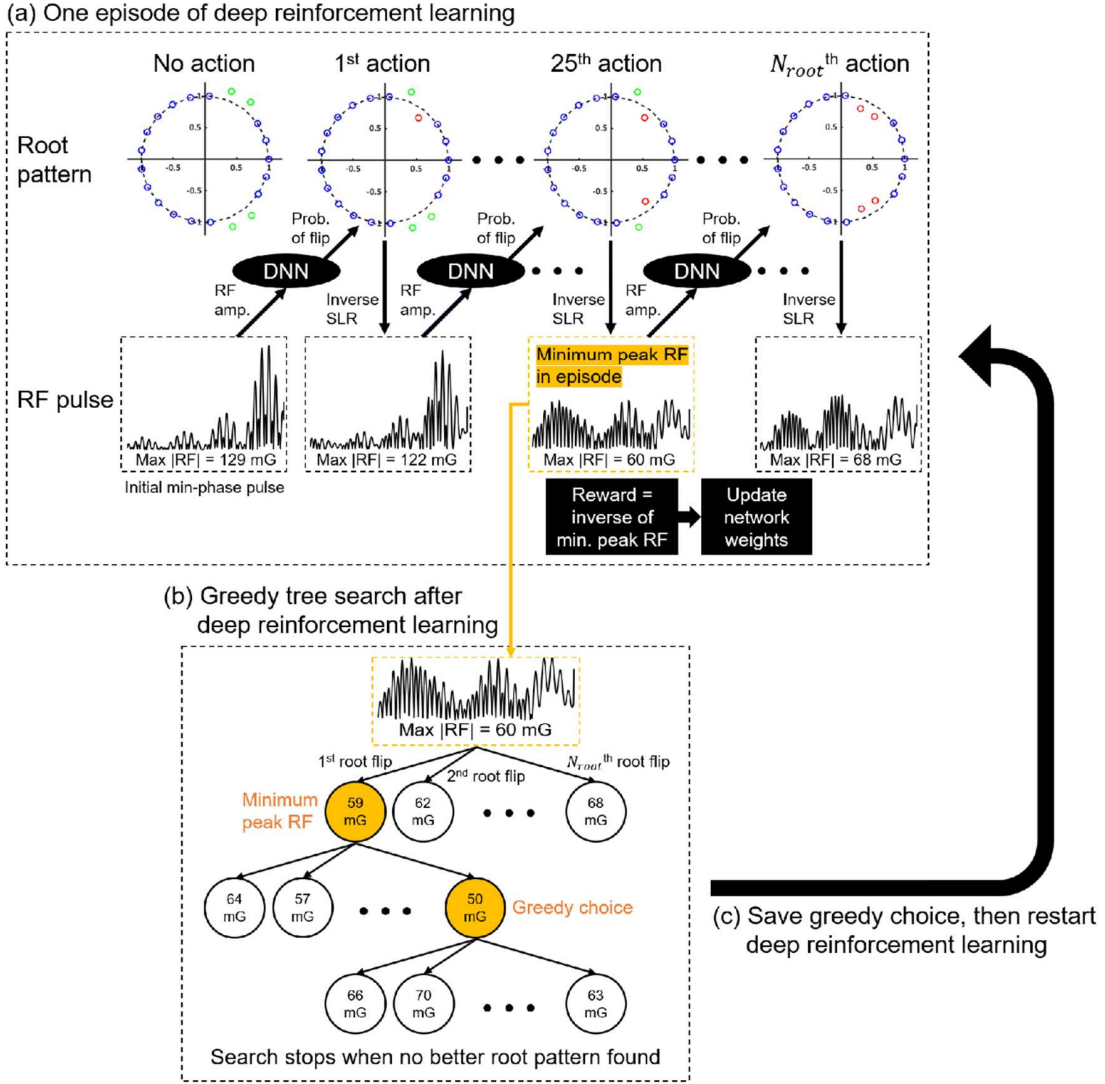

Fig. 2. Deep reinforcement learning and greedy tree search for DeepRF$_{SLR}$. (a) The deep reinforcement learning episode starts with the minimum phase RF pulse. Then, the DNN generates the probabilities of flipping to flip one root in each action. The corresponding RF pulse, which becomes the input to DNN, is determined by inverse SLR. Total $N_{root}$ number of RF pulses are generated in one episode, and the minimum peak RF among them is chosen to generate a reward to update the network weights. (b) The greedy tree search starts from the minimum peak RF in the deep reinforcement learning episode and grows the search tree with the minimum peak RF of each layer as the starting point. (c) When no better root pattern is found, the greedy tree search stops, and the greedy choice is saved. Then, DeepRF$_{SLR}$ iterates the whole process.

### C. Implementation Details of DeepRF$_{SLR}$

For implementation, two refocusing multiband RF pulses are designed for NB of 3 and 7 with the corresponding numbers of the eligible roots of 18 and 40, respectively. The TBW is set to be 6 with 1% stopband and passband ripple constraints. The centers of the two adjacent slices are spaced by six times of the slice thickness. The number of time points in RF is 512. All pulses are scaled to have the minimum duration for a given RF peak constraint of 200 mG.

The architecture of the DNN consists of an input layer, seven hidden layers, and an output layer. The input layer is composed of 512 neurons, and each hidden layer has 256 fully-connected neurons with leaky rectified linear units ($\alpha$ = 0.3) [32]. The output layer is a softmax layer [33] with $N_{root}$ neurons. The roots eligible for flipping (i.e., eligible roots) are defined as the roots within three times of bandwidth from the center of the passband in the top half of the unit circle [21]. The number of the eligible roots increases as the TBW and NB increases, and is not related to the number of time points [26]. For the roots in the bottom half, the roots in the top half are mirrored to enforce conjugate symmetry [28].

When performing the $N_{root}$ number of actions in each episode, we enforce that the same root is not flipped again during the episode. At the end of the episode, the reward is calculated as the inverse of minimum peak RF (i.e., $1/\min_{n=1,\ldots,N_{root}} RF_{peak}(n)$ where $RF_{peak}(n)$ refers to the peak RF after $n^{th}$ action). After estimating the reward, the DNN is trained with no discount rate [7], [30]. Adam optimizer, which is a first-order gradient-based optimization algorithm, is utilized for the optimization of the network weights with a learning rate of 0.0001 [34].

The DNN is implemented using TensorFlow [35]. The functions of the root-flipping method are adopted from the source codes available online [21], [28] and processed in MATLAB (MathWorks, Natick, Massachusetts, USA). In more detail, basic MATLAB functions for the RF pulse design are from the rf_tools software package (http://rsl.stanford.edu/research/software.html). The source codes of Sharma *et al*. (http://www.vuiis.vanderbilt.edu/~grissowa/) are slightly modified to enforce conjugate symmetry in the roots instead of anti-symmetry in the original work. All the modifications and comments are available at https://github.com/SNU-LIST/DeepRF_SLR. The computing environment is Intel(R) Core$^{TM}$ i7-7800X 3.50 GHz CPU, NVIDIA GeForce GTX 1080 Ti, and 64 GB memory.

Once the final refocusing pulse is designed, a phase-matched excitation pulse is generated using the work by Zun *et al*. [36] to complete a spin-echo sequence.

### D. Computer Simulations

The multiband RF with NB of 3 has 18 eligible roots for flipping and, therefore, has $2^{18}$ (= 262,144) binary combinations. For these combinations, an exhaustive search can find the optimal root pattern. On the other hands, the multiband RF with NB of 7 has a substantially larger number of combinations ($2^{40}$; approximately 1 trillion) and is not feasible to perform an exhaustive search. As a result, no optimum root pattern can be confirmed. For these differences, the RF designs for NB of 3 were compared for the computational time, reaching the optimum solution. Both the number of flipping and execution time was measured for all three methods (i.e., DeepRF$_{SLR}$, Monte-Carlo algorithm, and exhaustive search). Since the pulses are scaled to have the minimum duration for the peak RF constraint, the pulse duration was calculated from 8 executions of the DeepRF$_{SLR}$ and Monte-Carlo algorithms. For the design of RF with NB of 7, the DeepRF$_{SLR}$ and Monte-Carlo algorithms were terminated when the number of flipping reached a pre-defined repetition of 500,000. Then, the pulse duration was measured from 8 executions.

To demonstrate the advantages of DeepRF$_{SLR}$ for different values of TBW (6, 8, 10, and 12) and band gap (2, 4, 6, and 8), additional simulations were performed with the default simulation parameters of NB = 7, TBW = 6, band gap = 6 times of slice thickness, pre-defined repetition of 500,000 flips, and 5 executions.

### E. Phantom Experiments

To test the validity of the RF pulse in a scanner, a cylindrical phantom was scanned at a 3T Trio system (Siemens Medical Solutions, Erlangen, Germany) equipped with a 32-channel receiver head coil. The multiband pulses (NB of 3) designed by the DeepRF$_{SLR}$ and minimum-phase SLR algorithm [21] were compared. To visualize a slice profile, a sagittal plane was imaged after applying an excitation-refocusing pulse pair along the z-axis. Then, central 100 lines in the image were averaged along the phase-encoding direction to generate a slice profile plot. The scan parameters were as follows: repetition time = 500 ms ($T_1$ of the phantom = 100 ms), echo time = 29 ms, field of view = 25.6 × 25.6 cm$^2$, voxel size = 0.5 × 0.5 mm$^2$, thickness = 7 mm, readout bandwidth = 150 Hz/px, and scan time = 4 minutes and 16 seconds. The receive $B_1$ inhomogeneity effect was corrected by dividing the slice profile by the profile of a reference scan. The reference scan, which was a 3D GRE scan, had the following parameters: Repetition time = 40 ms, flip angle = 10°, echo time = 4.8 ms, field of view = 25.6 × 25.6 cm$^2$, voxel size = 0.5 × 0.5 × 5 mm$^3$, number of slices = 22, and scan time = 7 minutes and 30 seconds. The reference image was resampled to match the slice thickness.

To compare the experimental profiles with those of computer simulation, simulated slice profiles were calculated for the multiband excitation-refocusing pulse pairs of the DeepRF$_{SLR}$ and minimum-phase SLR algorithms. First, 1001 spins were placed equidistantly from -10 cm to 10 cm along the z-axis, and each spin was initialized to has a unit longitudinal magnetization. Then, the excitation and refocusing pulses with the slice selection and crusher gradients from the experiment were applied to the spins. The magnetization vectors of the spins were calculated by solving a discrete-time Bloch equation. Finally, the magnitudes of the transverse magnetization were obtained as the simulated slice profile. No $T_1$ and $T_2$ decays and transmit and receive $B_1$ inhomogeneities were considered.

## III. Results

The multiband refocusing RF pulses designed by the DeepRF$_{SLR}$, Monte-Carlo, and exhaustive search algorithm are shown in Fig. 3 for NB of 3. The results of the DeepRF$_{SLR}$ and exhaustive search reached the optimal root pattern and produced an RF pulse with the duration of 5.77 ms (Fig. 3a). However, the Monte-Carlo algorithm converged to the sub-optimum RF pulse which had the duration of 5.79 ms although the difference was small (Fig.3b). The number of root flipping and execution time to reach the optimal pattern were substantially shorter in DeepRF$_{SLR}$ (3,262 ± 3,100 flipping and 2.4 ± 2.2 minutes) than in the other two methods (exhaustive search: 262,144 flipping and 108 minutes; Monte-Carlo algorithm: not finding the optimal pattern for 300,000 flipping and 124 minutes; Table I). The failure to converge to the optimum result in the Monte-Carlo algorithm was due to the generation of duplicate patterns (see Discussion). In Fig. 3c, the pulse durations are plotted over the number of flipping for the results of the DeepRF$_{SLR}$ (red line) and Monte-Carlo (green line) algorithms. DeepRF$_{SLR}$ shows a much faster convergence to the optimum pattern. The root patterns are shown in Fig. 3d and 3e for DeepRF$_{SLR}$ (and the exhaustive search) and the Monte-Carlo algorithm, respectively. The eligible roots for flipping are denoted as green dots and the flipped roots as red dots. The root pattern of the minimum-phase SLR method is shown in Supplementary Fig. S1 for comparison.

TABLE I
AVERAGE NUMBERS OF FLIPPING AND EXECUTION TIMES TO REACH THE OPTIMAL ROOT PATTERN FOR THE MULTIBAND REFOCUSING RF PULSES WITH NB OF 3 FOR THE EXHAUSTIVE SEARCH, MONTE-CARLO, AND DEEPRF$_{SLR}$ ALGORITHMS. DEEPRF$_{SLR}$ REACHED THE OPTIMAL PATTERN FASTER THAN THE OTHER TWO METHODS, REQUIRING THE SUBSTANTIALLY SMALLER NUMBER OF FLIPPING AND SHORTER EXECUTION TIME.

|  | (Average) Number of flipping | (Average) Execution time |
|---|---|---|
| Exhaustive search | 262,144 | 108 minutes |
| Monte-Carlo algorithm | > 300,000 | > 124 minutes |
| DeepRF$_{SLR}$ | 3,262 ± 3,100 | 2.4 ± 2.2 minutes |

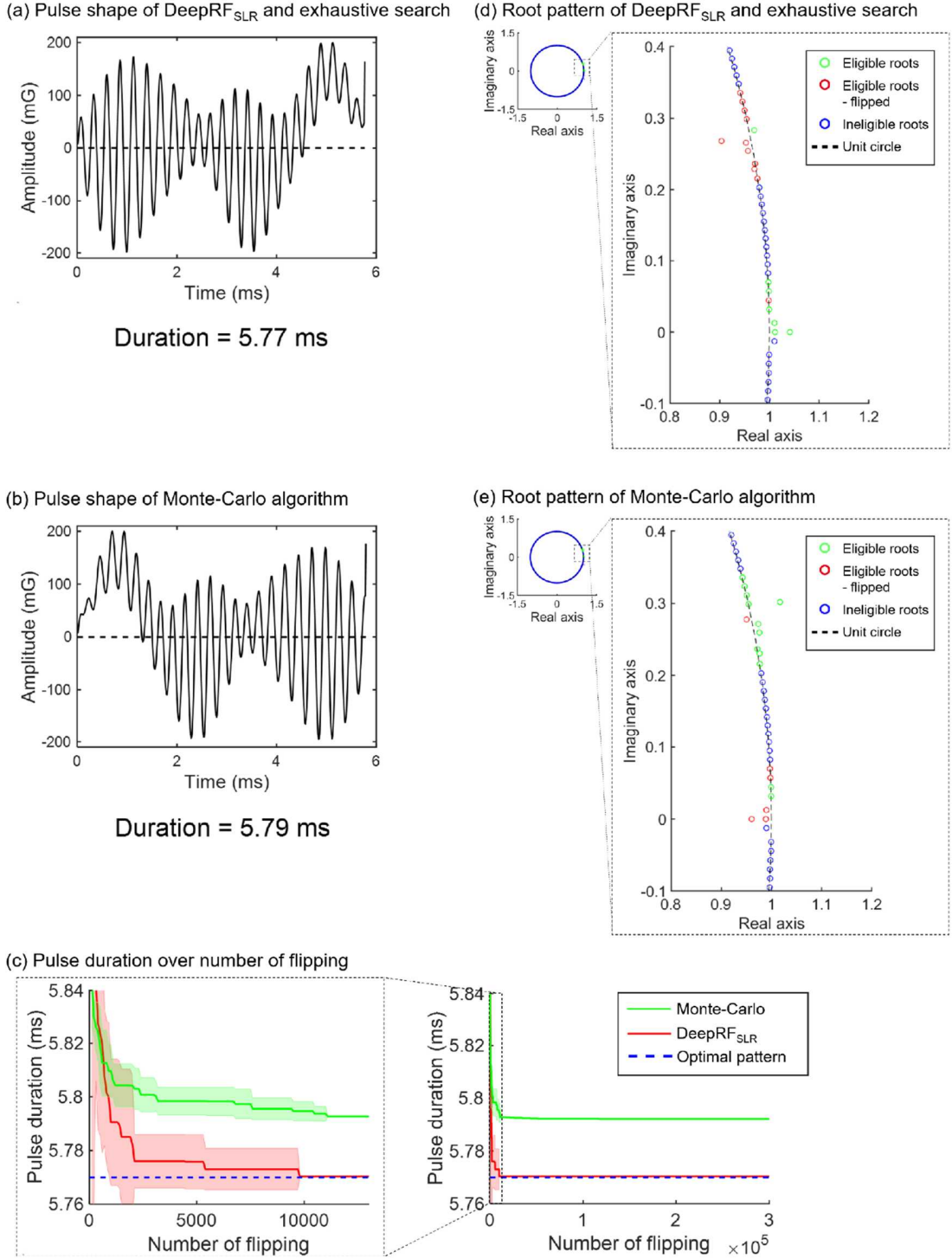

Fig. 3. Multiband refocusing RF pulses for NB of 3. (a) The RF pulse shape of the optimal root pattern from DeepRF$_{SLR}$ and exhaustive search. The duration was 5.77 ms. (b) The RF pulse shape from the Monte-Carlo algorithm. The duration was 5.79 ms. (c) The RF pulse durations over the number of flipping for the DeepRF$_{SLR}$ and Monte-Carlo algorithms. All eight executions of DeepRF$_{SLR}$ (red) converged to the optimal pattern within 10,000 flipping, whereas all the executions of the Monte-Carlo algorithm (green) did not converge to the optimal pattern for 300,000 flipping. The shaded area around the solid line reports one standard deviation. (d) The optimal root pattern found by the DeepRF$_{SLR}$ and exhaustive search algorithms. (e) The root pattern of the Monte-Carlo algorithm. Green dots are unflipped eligible roots (i.e., same as minimum-phase RF) whereas red dots are flipped eligible roots. Ineligible roots are blue dots.

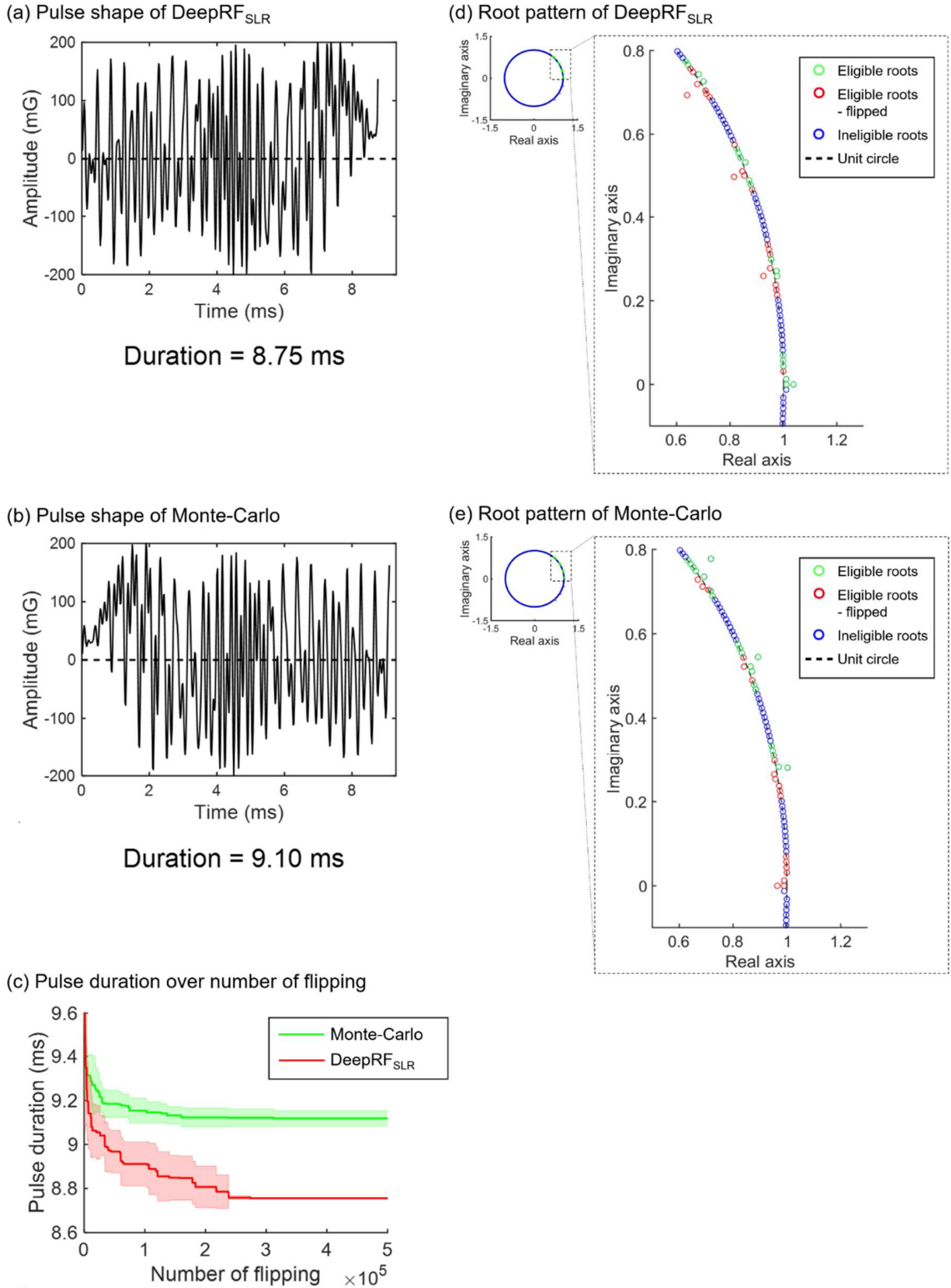

Fig. 4. The multiband refocusing RF pulses for NB of 7. (a) The RF pulse shape from DeepRF$_{SLR}$. The duration was 8.75 ms. (b) The RF pulse shape from the Monte-Carlo algorithm. The duration was 9.10 ms. (c) The RF pulse durations over the number of flipping for the DeepRF$_{SLR}$ and Monte-Carlo algorithms. The eight executions of the DeepRF$_{SLR}$ (red) and Monte-Carlo (green) algorithms are plotted. The shaded area around the solid line reports one standard deviation. The root patterns found by DeepRF$_{SLR}$ (d) and Monte-Carlo (e). Green dots are unflipped eligible roots (i.e., same as minimum-phase RF) whereas red dots are flipped eligible roots. Ineligible roots are blue dots.

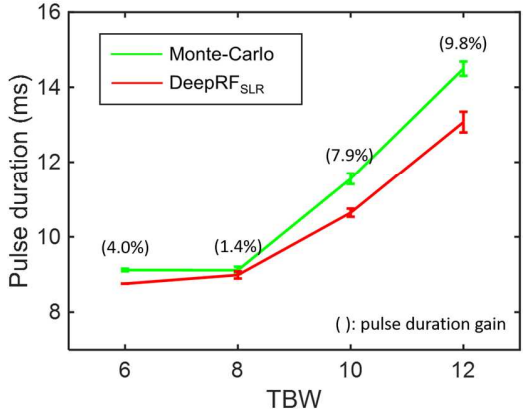 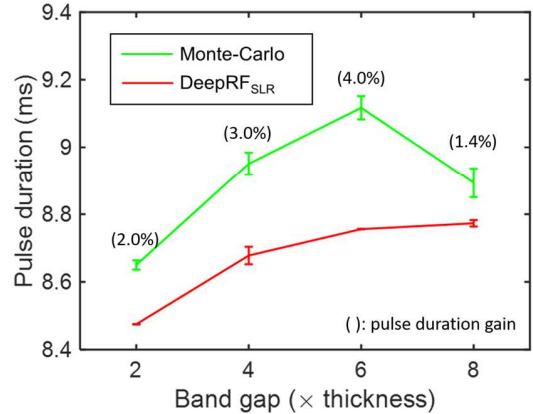

Fig. 5. The RF pulse durations for different TBWs (6, 8, 10, and 12) and band gaps (2, 4, 6, and 8). The results of DeepRF$_{SLR}$ report shorter durations than those of Monte-Calrlo in all cases. (a) The pulse duration gain, which is defined as the ratio of the pulse duration of Monte-Carlo-designed RF and that of DeepRF$_{SLR}$-designed RF, increases as TBW increases from 8 to 12. (b) The band gap shows no clear relationoiship to the pulse duration gain.

The RF pulses with NB of 7 designed by the DeepRF$_{SLR}$ and Monte-Carlo algorithms are shown in Fig. 4. The algorithms stopped after 500,000 flipping. No exhaustive search was performed since the possible root combinations are too many. The RF pulse by DeepRF$_{SLR}$ had a shorter duration than that of the Monte-Carlo algorithm (8.75 ms for DeepRF$_{SLR}$ vs. 9.10 ms for Monte-Carlo; Fig. 4a&b). The pulse duration plotted over the number of flipping is shown in Fig. 4c, revealing DeepRF$_{SLR}$ finds a better solution. When plotting the pulse duration over the execution time (Supplementary Fig. S2), DeepRF$_{SLR}$ still shows the superior results. The root patterns of the two methods are different (Fig. 4d&e). These results demonstrate that DeepRF$_{SLR}$ optimizes the multiband pulse more efficiently than the Monte-Carlo algorithm for a large NB and the difference (0.35 ms) can be meaningful.

When tested for different TBWs, the DeepRF$_{SLR}$-designed RF pulses demonstrated larger pulse duration gains (i.e. shorter pulse duration in the ratio) than those of the Monte-Carlo-designed RF pulses for a higher TBW than 8 (Fig. 5a). The gain increased from 4% when TBW was 6 (0.35 ms difference in the pulse duration) to 9.8% when TBW was 12 (1.43 ms difference), demonstrating clear advantages of DeepRF$_{SLR}$ for difficult problems with large numbers of eligible roots. For different band gaps, no clear trend was observed but the DeepRF$_{SLR}$ results offered shorter durations over those from the Monte-Carlo algorithms (Fig. 5b).

For the MRI experiments, the minimum-phase SLR excitation-refocusing pulse pair and DeepRF$_{SLR}$ pulse pair for NB of 3 are shown in Fig. 6. The DeepRF$_{SLR}$ excitation pulse had 2.3 times shorter duration (3.04 ms) than the duration of the minimum-phase excitation pulse (6.90 ms). The DeepRF$_{SLR}$ refocusing pulse had 2.1 times shorter duration (5.77 ms) than that of the minimum-phase refocusing pulse (12.31 ms).

When these pulses were applied for the phantom scan, they successfully generated multiband images (Fig. 7). The measured slice profile of DeepRF$_{SLR}$ (solid line) was similar to that of the minimum-phase SLR (dashed line; Fig. 7d) although slight distortions in the passband region (arrows in Fig. 7d) were observed in the DeepRF$_{SLR}$ profile (see Discussion). The simulated slice profiles of the two methods resulted in almost identical profiles (Fig. 7c).

## IV. Discussion

In this paper, we proposed a novel RF pulse design method, DeepRF$_{SLR}$, that optimized the RF pulse using deep reinforcement learning. In particular, DeepRF$_{SLR}$ optimized a multiband RF pulse using the SLR root flipping approach, generating a reduced peak amplitude or, equivalently, a shorter pulse duration. For NB of 3, the execution time to reach the optimal pattern was dramatically reduced for DeepRF$_{SLR}$. For NB of 7, DeepRF$_{SLR}$ found a shorter duration RF than the Monte-Carlo algorithm. In the phantom experiment, the slice profile using the DeepRF$_{SLR}$ excitation-refocusing pulse pair was approximately equivalent to that from the minimum-phase pulse pair while reducing the RF duration by a factor of 2.1 and more.

In DeepRF$_{SLR}$, deep reinforcement learning was combined with the greedy tree search to perform an efficient optimization. When deep reinforcement learning was applied alone, the performance was degraded (see Supplementary Fig. S3). The greedy tree search is known to be sensitive to the starting point and does not find the optimal solution in a complex problem [29]. The combination of these two types of algorithms, however, has successfully performed the optimization and has been applied in other applications [37], [38]. Deep reinforcement learning has also been combined with other algorithms such as beam search [39] and random search [40] to improve the search results. One intuitive interpretation for the functions of the two methods is that deep reinforcement learning generates a better seed point for the greedy tree search. Alternatively, one can consider deep reinforcement learning performing exploration whereas the greedy tree search performing exploitation. Hence, the two methods conduct complementary roles in finding a solution. When combining the two algorithms, finding an approximate learning rate of deep reinforcement learning is important. Since the greedy search starts from the seed point given by the agent of deep reinforcement learning, if the learning rate is too high, the agent may stick to a trivial solution and lose the opportunities to generate good seed points for the greedy search. On the other hand, if the learning rate is too low, the neural network weights are updated too slowly and generating good seed points takes long time [7].

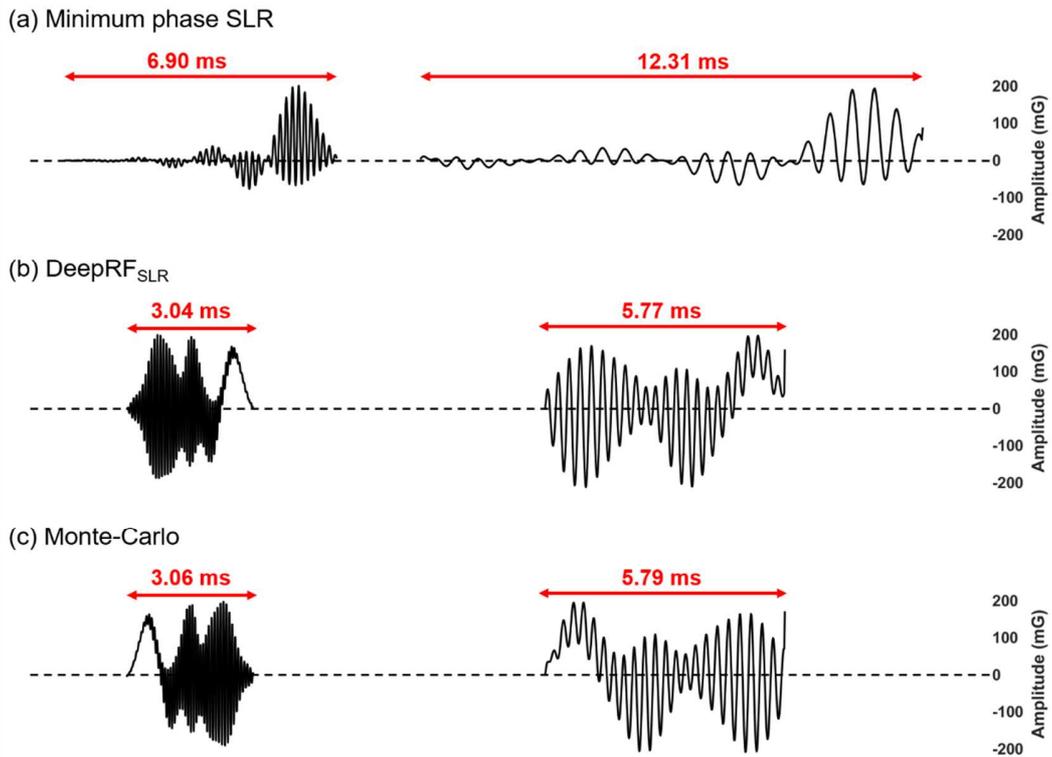

Fig. 6. Comparison of the minimum-phase SLR, DeepRF$_{SLR}$, Monte-Carlo excitation-refocusing pulse pairs for NB of 3. The DeepRF$_{SLR}$ excitation pulse has 2.3 times shorter duration and DeepRF$_{SLR}$ refocusing pulse has 2.1 times shorter duration than that of the minimum-phase pulse pair.

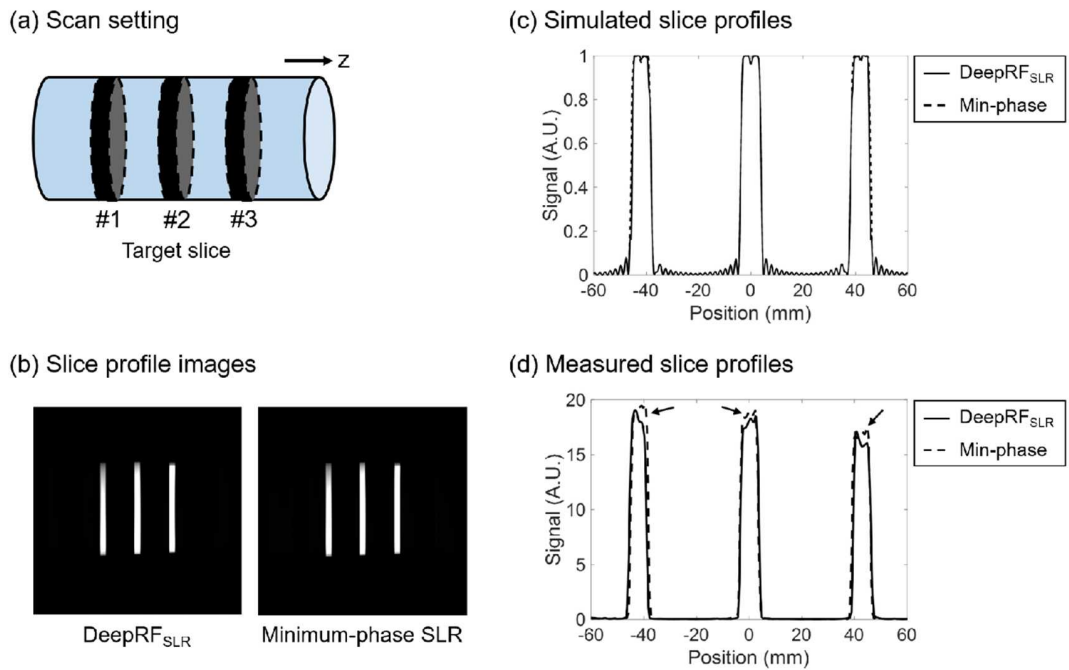

Fig. 7. Experimental results. (a) The phantom scan setting. (b) The sagittal images of the phantom acquired with DeepRF$_{SLR}$ and minimum-phase SLR. (c) The simulated slice profiles (solid line: DeepRF$_{SLR}$ and dashed line: minimum-phase SLR). (d) The measured slice profiles (solid line: DeepRF$_{SLR}$ and dashed line: minimum-phase SLR).

The measured slice profile of DeepRF$_{SLR}$ in Fig. 7d reported slight distortions. To understand the origin of the distortions, we investigated the effects of the spatial resolution. The experimental results, that are summarized in Supplementary Fig. S4, suggested that the spatial resolution affected the measurement but was not the primary source of the distortions. Further research is necessary to understand the origin of the distortions.

When comparing the pulse shapes from the DeepRF$_{SLR}$ and Monte-Carlo methods, the two roughly showed a mirrored-shape (Fig. 3a&b and Fig. 4a&b). Since the time-reversal of a pulse is achieved by flipping all the roots, one may generate a DeepRF$_{SLR}$-like RF pattern by flipping all the eligible roots of the Monte-Carlo RF.

For the RF pulse with NB of 3, the Monte-Carlo algorithm failed to find the optimal pattern even after 300,000 flips, which are larger than the number of flips for the exhaustive search. This can be explained by the Monte-Carlo algorithm generating a random number in each trial with no memory [21]. As a result, it produces the same pattern multiple times, degrading the performance.

When designing the RF pulse with NB of 7, the execution time reaching the fixed number of flipping (= 500,000) was longer in DeepRF$_{SLR}$ than in the Monte-Carlo algorithm (see Supplementary Table SI). This is due to the use of the graphical processing unit (e.g., additional matrix computation and data transfer time) in DeepRF$_{SLR}$.

In Supplementary Fig. S5, the results of different peak RF constraints are summarized, reporting consistent benefits of DeepRF$_{SLR}$ for different constraints.

Although we demonstrated the multiband refocusing pulse designs using DeepRF$_{SLR}$, DeepRF$_{SLR}$ can be used to generate a single-band RF pulse. The method may be useful in designing a single-band pulse with a large TBW.

Despite the efficiency of DeepRF$_{SLR}$, the computational time ranged from a few minutes to hours. This indicates that one cannot design a multiband RF in real-time. To increase efficiency, one can implement a multi-agent system [41]. Alternatively, one may utilize transfer learning from a ready-trained network to reduce the training time [42]. The computational efficiency of deep learning-designed RF will find more applications in the future (e.g., $B_1$ inhomogeneity-robust multiband pulse [43], [44] or parallel transmission multiband pulse designs [45]).

## V. CONCLUSION

In conclusion, a novel deep reinforcement learning framework, DeepRF$_{SLR}$, that optimizes the multiband RF pulse via SLR root flipping is introduced. The results demonstrate that the duration of the RF can be reduced when compared to the Monte-Carlo algorithm-designed RF or minimum-phase RF. Our approach suggests a new way of designing an RF by applying a machine learning algorithm, demonstrating a "machine-designed" MRI sequence.

# Supplementary Materials for "Deep Reinforcement Learning Designed Shinnar-Le Roux RF Pulse using Root-Flipping: DeepRF$_{SLR}$"

Dongmyung Shin, Sooyeon Ji, Doohee Lee, Jieun Lee, Se-Hong Oh, and Jongho Lee

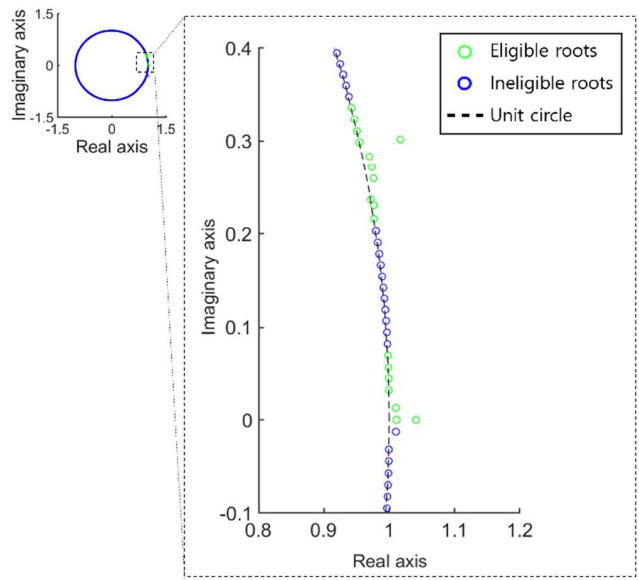 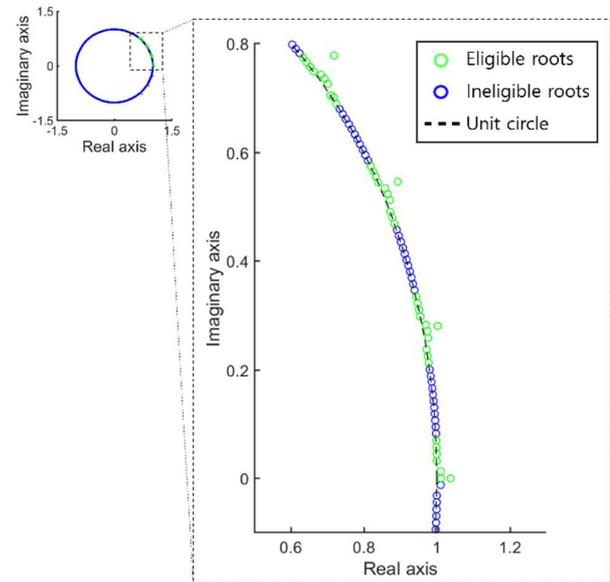

Fig. S1. Root patterns of the minimum-phase SLR method for NB of 3 and 7 RF pulses.

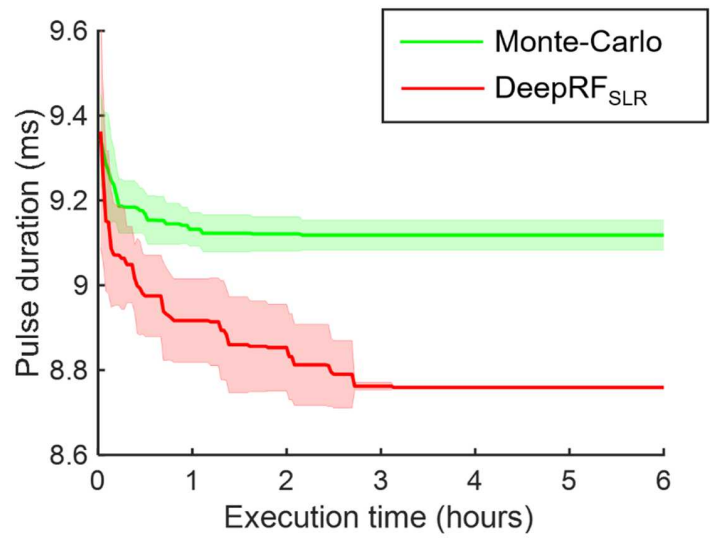

Fig. S2. Pulse durations over the execution time for NB of 7 RF pulses. Both DeepRF$_{SLR}$ and Monte-Carlo results are plotted.

TABLE SI
MEAN EXECUTION TIMES FROM THE 8 EXECUTIONS OF THE MONTE-CARLO ALGORITHM AND DEEPRF$_{SLR}$ REACHING 500,000 FLIPPING WHEN DESIGNING NB OF 7 RF PULSES.

|  | Execution time |
|---|---|
| Monte-Carlo | $207 \pm 1$ minutes |
| DeepRF$_{SLR}$ | $345 \pm 9$ minutes |

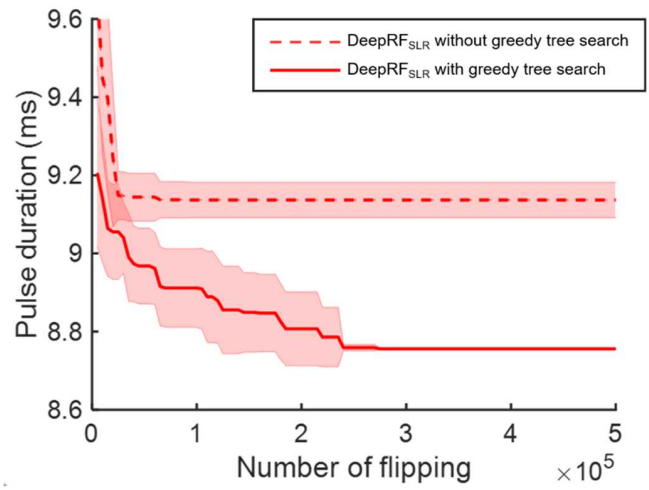

Fig. S3. Pulse durations over the number of flipping for NB of 7 RF pulses designed by DeepRF$_{SLR}$ with and without the greedy tree search.

(a) Slice profile images (thickness = 12 mm)

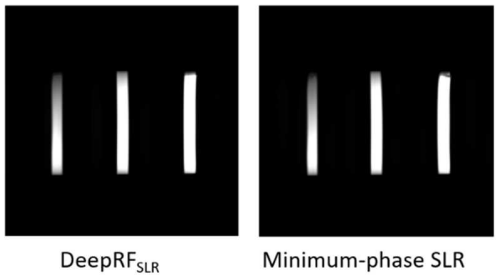

DeepRF$_{SLR}$      Minimum-phase SLR

(c) Center slice images (thickness = 50 mm)

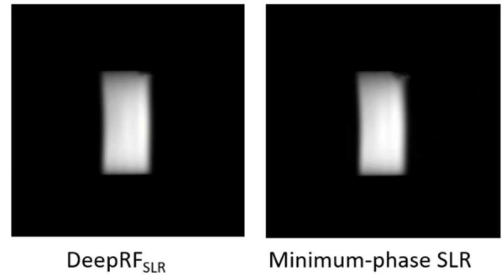

DeepRF$_{SLR}$      Minimum-phase SLR

(b) Measured slice profiles after $B_1^-$ correction (thickness = 12 mm)

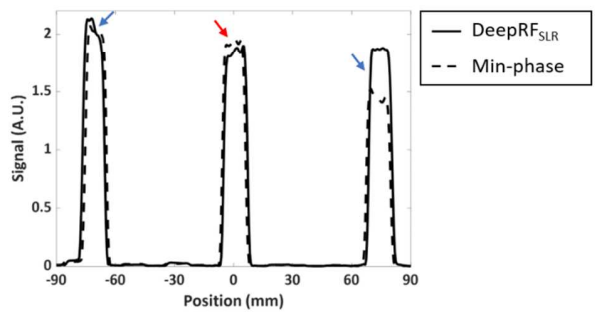

(d) Center slice profiles after $B_1^-$ correction (thickness = 50 mm)

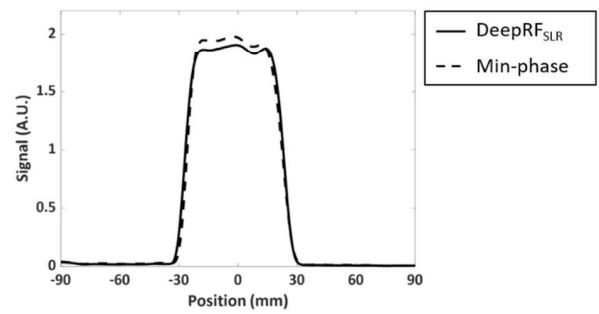

Fig. S4. Experimental results with thicker slices (left: 12 mm; right: 50 mm). (a) The sagittal images of the phantom acquired with the DeepRF$_{SLR}$ and minimum-phase SLR RF pulses with the slice thickness of 12 mm. (b) DeepRF$_{SLR}$ profile shows similar or better results (blue arrows) despite small distortions in the center slice (red arrow). (c) Experiments performed with the thickness of 50 mm. Only the center slice was imaged due to the limited length of the phantom. (d) Distortions in the center slice still exist for the thickness of 50 mm. These results suggest that the spatial resolution affected the measured slice profiles. However, it may not be the primary source of the distortions.

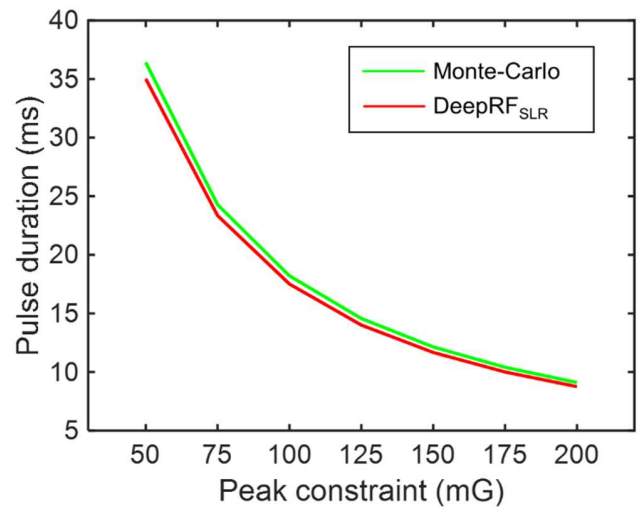

Fig. S5. Simulation results for different peak constraints. A lower peak constraint shows a longer pulse duration but the pulse duration gains between the two designs were the same (= 4%).